\documentclass[a4paper,10pt, twocolumn]{article}
\usepackage[utf8]{inputenc}
\usepackage{lipsum} 
\usepackage{baskervald}
\usepackage{amssymb,amsfonts}
\usepackage{amsmath,amsthm}
\usepackage[all,arc, cmtip]{xy}
\usepackage{enumerate}
\usepackage{caption}
\usepackage{subcaption}
\usepackage{tikz}
\usetikzlibrary{arrows,decorations.pathmorphing,backgrounds,positioning,fit,petri}
\usetikzlibrary{decorations.pathreplacing,calligraphy}
\usepackage{paralist}
\usepackage{booktabs} 
\usepackage{array} 
\usepackage{mathrsfs}
\usepackage{url}
\usepackage{color}
\usepackage{cancel}
\usepackage{todonotes} 
\usepackage{hyperref} 
\usepackage{mathtools}
\usepackage{multirow}
\usepackage[mathscr]{euscript}
\usepackage[nameinlink,capitalise,noabbrev]{cleveref}
\usepackage[shortcuts]{extdash}

\usepackage[T1]{fontenc}
\usepackage{authblk}

\newcommand{\Feat}{\mathrm{Feat}}

\usepackage{tikz-cd}

\newtheorem{thm}{Theorem}[section]
\newtheorem*{thm*}{Theorem}

\newtheorem{lem}[thm]{Lemma}

\theoremstyle{definition}

\newtheorem{defn}[thm]{Definition}
\newtheorem*{defn*}{Definition}

\newtheorem{rem}[thm]{Remark}

\theoremstyle{remark}

\newcommand{\TS}{\mathrm{TS}}
\newcommand{\RR}{\mathbb{R}}

\makeatletter
\let\c@equation\c@thm
\makeatother
\numberwithin{equation}{section}
\title{Graph Neural Networks and Time Series as Directed Graphs for Quality Recognition}
\author[1]{Angelica Simonetti\thanks{a.simonetti@lancaster.ac.uk}}
\author[2]{Ferdinando Zanchetta\thanks{ferdinando.zanchett2@unibo.it\\ The authors have contributed equally to this paper.}}
\affil[1]{Department of Mathematics and Statistics, University of Lancaster}
\affil[2]{Fabit, University of Bologna}

\tikzstyle{nd1}=[circle,draw=blue!50,fill=blue!10!,thick,
inner sep=0pt,minimum size=3mm, rounded corners=1ex]
\tikzstyle{nd2}=[circle,draw=blue!50,fill=blue!10!,thick,
inner sep=0pt,minimum size=3mm, rounded corners=1ex]
\tikzstyle{nd3}=[circle,draw=blue!50,fill=blue!10!,thick,
inner sep=0pt,minimum size=3mm, rounded corners=1ex]

\tikzstyle{inout}=[rectangle,draw=black!50,fill=black!10,thick,
inner sep=0pt, minimum width=12mm, minimum height=5mm, rounded corners=1ex]
\tikzstyle{layer}=[rectangle,draw=blue!50,fill=blue!10,thick,
inner sep=0pt, minimum width=20mm, minimum height=5mm, rounded corners=1ex]
\tikzstyle{main}=[rectangle,draw=black!50!green,fill=black!20!green!15,thick,
inner sep=0pt, minimum width=20mm, minimum height=5mm, rounded corners=1ex]
\tikzstyle{bn}=[circle,draw=red!50,fill=red!10,thick,
inner sep=0pt,minimum size=8mm, rounded corners=1ex]
\tikzstyle{bg}=[fill=black!5!yellow!20, minimum width=25mm, minimum height=20mm, rounded corners=1ex]
\tikzstyle{bg2}=[fill=black!5!green!20, minimum width=25mm, minimum height=20mm, rounded corners=1ex]
\tikzstyle{bg3}=[fill=black!5!yellow!20, minimum width=25mm, minimum height=29mm, rounded corners=1ex]
\tikzstyle{times}=[rectangle,draw=black!80!yellow!60, fill=black!20!yellow!20, minimum width=5mm, minimum height=5mm, rounded corners=1ex]
\tikzstyle{times2}=[rectangle,draw=black!80!green!60, fill=black!20!green!20, minimum width=5mm, minimum height=5mm, rounded corners=1ex]

\begin{document}
\maketitle
\begin{abstract}
Graph Neural Networks (GNNs) are becoming central in the study of time series, coupled with existing algorithms as Temporal Convolutional Networks and Recurrent Neural Networks.
In this paper, we see time series themselves as directed graphs, so that their topology encodes time dependencies and we start to explore the effectiveness of GNNs architectures on them. We develop two distinct Geometric Deep Learning models, a supervised classifier and an autoencoder-like model for signal reconstruction. We apply these models on a quality recognition problem.
\end{abstract}

\section{Temporal convolutional Networks}
Convolutional neural networks (CNNs, see \cite{lbbh}, \cite{LeNet}, \cite{lecun2010mnist}, \cite{kr-imagenet}, \cite{LeCun2015}) are deep learning algorithms employing so called \emph{convolutional layers}: these are layers that are meant to be applied on grid-like data, e.g. images. For data organized in sequences, 1d CNNs were developed (\cite{1DCNN19}, \cite{Kiranyaz2019}) and, more recently, TCNs have become popular in the study of time series  (see \cite{Bai2018} and the references therein). Throughout the paper, $\TS(r,m)$ will denote the set of multivariate time series with $m$ channels and length $r$ in the temporal dimension. Given $\textbf{x}\in\TS(r,m)$ we will denote as $\textbf{x}(i)_j$ (or simply as $\textbf{x}_{ij}$ when no confusion is possible), for $i=1,...,r$ and $j=1,...,m$, the $j$ the coordinate of the vector $\textbf{x}(i)\in\RR^m$. For a given natural number $n$, we shall denote as $[n]$ the ordered set $(1,...,n)$. Now, recall that given a filter $K\in\RR^f$, we can define a one-channel, one-dimensional (1D) convolution as an operator
\begin{align*}
    &\mathrm{conv1D}:\TS(r,1)\rightarrow\TS(l,1) \\
    &\mathrm{conv1D}(\textbf{x})_j=\sum_{i=1}^fK_i\textbf{x}_{\alpha(j,i)}+b_j
\end{align*}
where $\alpha(j,-):[f]\rightarrow\mathbb{Z}$ are injective index functions, $\textbf{x}_i:=0$ if $i\notin [r]$ and $b\in\mathbb{R}^l$ is a bias vector.
The numbers $K_i$ are called the \emph{parameters} or \emph{weights} of the convolution. The most commonly used index functions are of the form $\alpha(j,i)=(n+d\cdot i)+j$ for some integers $n,d$. As a consequence, from now on we shall assume that the one dimensional convolutions we consider have this form. If, $\alpha(j,i)\leq j$ for all $i,j$ then the convolution is said to be \emph{causal} as it will look only 'backward'. These are the building blocks of TCNs, that are CNNs where only causal convolutions appear. If $|d|>1$, the convolution is called \emph{dilated}. 
One could define multi-channel (i.e. handling multivariate time series), 1D convolutions in two steps. First, we define convolutions taking a multivariate time series to an univariate time series as operators $\mathrm{conv}:\TS(r,n)\rightarrow\TS(l,1)$ as $\mathrm{conv}(\textbf{x})_i=\sum_{j=1}^n\mathrm{conv1D}_j(\textbf{x}(-)_{(j)})$ where $\mathrm{conv1D}_j$ are one-channel, one-dimensional convolutions. Then we can define 1D convolutions transforming multivariate time series into multivariate time series as operators $\mathrm{conv}:\TS(r,n)\rightarrow\TS(l,m)$ that are multi-channel 1D convolutions when co-restricted at each non temporal dimension of the output.
The usefulness of TCNs in the context of time series arises from the fact that causal convolutions by design are able to exploit temporal dependencies, while not suffering from some of the algorithmic problems of RNNs such as LSTMs: for example they appear to be faster to train and more scalable (see \cite{Bai2018} for a discussion).

\begin{figure*}[t]
\vspace*{-0.5cm}
\begin{center}
\begin{tikzpicture}[scale=0.46]
\node at (0,0) [nd1] (p11) {};
\node at (1*3,0) [nd1] (p12) {};
\node at (2*3,0) [nd1] (p13) {};
\node at (3*3,0) [nd1] (p14) {};
\node at (4*3,0) [nd1] (p15) {};
\node at (5*3,0) [nd1] (p16) {};
\node at (6*3,0) [nd1] (p17) {};
\node at (7*3,0) [nd1] (p18) {};
\node at (8*3,0) [nd1] (p19) {};

\draw [->] (p11.east) -- (p12.west);
\draw [->] (p12.east) -- (p13.west);
\draw [->] (p13.east) -- (p14.west);
\draw [->] (p14.east) -- (p15.west);
\draw [->] (p15.east) -- (p16.west);
\draw [->] (p16.east) -- (p17.west);
\draw [->] (p17.east) -- (p18.west);
\draw [->] (p18.east) -- (p19.west);

\node at (0,1) [nd2] (p21) {};
\node at (1*3,1) [nd2] (p22) {};
\node at (2*3,1) [nd2] (p23) {};
\node at (3*3,1) [nd2] (p24) {};
\node at (4*3,1) [nd2] (p25) {};
\node at (5*3,1) [nd2] (p26) {};
\node at (6*3,1) [nd2] (p27) {};
\node at (7*3,1) [nd2] (p28) {};
\node at (8*3,1) [nd2] (p29) {};

\draw [->] (p21.east) -- (p22.west);
\draw [->] (p22.east) -- (p23.west);
\draw [->] (p23.east) -- (p24.west);
\draw [->] (p24.east) -- (p25.west);
\draw [->] (p25.east) -- (p26.west);
\draw [->] (p26.east) -- (p27.west);
\draw [->] (p27.east) -- (p28.west);
\draw [->] (p28.east) -- (p29.west);

\node at (8*3,2) {$\vdots$};
\node at (0,2) {$\vdots$};

\node at (-1*2.5,3.2) [nd3] (e0) {$\ v\ $};
\node at (0,3) [nd3] (e1) {};
\node at (1*3,3) [nd3] (e2) {};
\node at (2*3,3) [nd3] (e3) {};
\node at (3*3,3) [nd3] (e4) {};
\node at (4*3,3) [nd3] (e5) {};
\node at (5*3,3) [nd3] (e6) {};
\node at (6*3,3) [nd3] (e7) {};
\node at (7*3,3) [nd3] (e8) {};
\node at (8*3,3) [nd3] (e9) {};

\draw [->] (e1.east) -- (e2.west);
\draw [->] (e2.east) -- (e3.west);
\draw [->] (e3.east) -- (e4.west);
\draw [->] (e4.east) -- (e5.west);
\draw [->] (e5.east) -- (e6.west);
\draw [->] (e6.east) -- (e7.west);
\draw [->] (e7.east) -- (e8.west);
\draw [->] (e8.east) -- (e9.west);

\draw[->, decorate,decoration=snake] (e1.north west) -- (e0.east);

\path [->] (e1.north east) edge [bend left=15]  (e5.north west);
\path [->] (e1.north east) edge [bend left=6]  (e9.north west);
\draw [<->] (e1.south) edge [bend right=30] (p21.north);
\path [<->] (e1.south west) edge [bend right=30] (p11.north west);
\draw [->] (e1.south east) -- (p22.north west);
\draw [->] (e1.south east) -- (p12.north west);
\path [->] (e1.south east) edge [bend left=2]  (p25.north west);
\path [->] (e1.south east) edge [bend left=1]  (p29.north west);
\path [->] (e1.south east) edge [bend left=5]  (p15.north west);
\path [->] (e1.south east) edge [bend left=7.5]  (p19.north west);
\end{tikzpicture}
\end{center}
\caption{\footnotesize{One possible structure of a time-digraph, as described in Example 1. Here only adjacent connections and all the connections for node $v$ are shown, and $d=4$}}\label{tgraph}
\end{figure*}
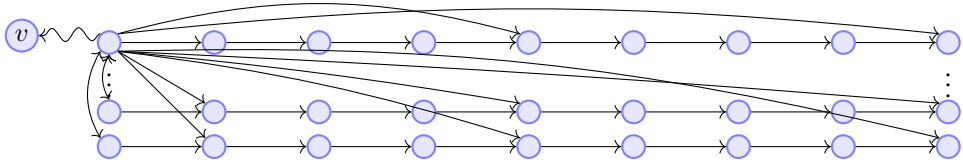
\section{Time series as Directed Graphs}
\subsection{Generalities.}
\begin{defn}
    A \emph{directed graph (digraph)} $G$ is the datum $G=(V_G,E_G,h_G,t_G)$ of two sets $V_G$ (the \emph{set of vertices}), $E_G$ (the \emph{set of edges}) and two functions $h_G,t_G:E_G\rightarrow V_G$ associating to each edge $e$ its \emph{head} $h_G(e)$ and its \emph{tail} $t_G(e)$ respectively. A morphism $\varphi:G\rightarrow H$ between two directed digraphs $G$ and $H$ is the datum of two functions $\varphi_V:V_G\rightarrow V_H$, $\varphi_E:E_G\rightarrow E_H$ such that $h_H\circ\varphi_E=\varphi_V\circ h_G$ and $t_H\circ\varphi_E=\varphi_V\circ t_G$
\end{defn}
From now on, for simplicity we will assume that our digraphs have at most one edge connecting two different nodes (for each direction) and at most one self loop for each node. In this case, given a digraph $G=(V_G,E_G,h_G,t_G)$ and an ordering of the vertices $(v_i)_{i\in [|V_G|]}$, we can define the \emph{adjacency matrix of $G$} as the matrix defined by $A_{ij} = 1$ if there is there exists an edge having as tail $v_i$ and head $v_j$ or $A_{ij} = 0$ otherwise. If the adjacency matrix of a graph is symmetric, we say that our graph is \emph{undirected}. We can assign \emph{weights} to the edges of a graph by letting the entries of the adjacency matrix to be arbitrary real numbers. When speaking about weighted digraphs, we shall always assume that there is an edge having as tail $v_i$ and head $v_j$ if and only if $A_{ij} \neq 0$. We speak in this case of \emph{weighted digraphs}. 
\begin{defn}
    A \emph{digraph with features of dimension $n$} is the datum $(G,F_G=(h_{v})_{v\in V_G})$ of a (weighted or unweighted) digraph $G$ and vectors $h_v\in\RR^n$ of \emph{node features} for each vertex $v\in V_G$. For a given digraph $G$, we shall denote as $\Feat(G,n)$ the set of all digraphs with features of dimension $n$ having $G$ as underlying digraph.
\end{defn}
Real-world graph-structured datasets usually come in form of one or more digraphs with features. Given a digraph with features $(G,F_G=(h_{v})_{v\in V_G})$ and a digraph morphism $\varphi:H\rightarrow G$, we can pullback the features of $G$ to obtain a graph with features $(H,\varphi^\ast F_G=(h_{\varphi_V(v)})_{v\in V_H})$. This defines a function $\varphi^\ast:\Feat(G,n)\rightarrow\Feat(H,n)$. Graph Neural Networks (GNNs, \cite{GNNSurvey19}) are models that are used on graph-structured data using as building blocks the so called \emph{graph convolutions} (\cite{Bronstein2016GeometricDL}, \cite{Fioresi2023DeepLA}): given a graph, they update each node feature vector combining the information contained in the feature vectors of adjacent nodes. In general, a graph convolution is a function $\mathrm{gconv}:\Feat(G,n)\rightarrow\Feat(G,m)$ that is permutation invariant in a sense that we shall make precise below. graph convolutions update node features of a digraph using a \emph{message passing mechanism} that can be written in the following general form 
\begin{equation}\label{eq1}
 h_{v_i}'=\sigma(\psi(h_{v_i}, \oplus_{v_j\in N^{\alpha}(v_i)}\varphi(h_{v_i},h_{v_j},A_{ij},A_{ji}))
 \end{equation}
where $\sigma$ is an activation function, $\alpha\in\lbrace h,t,u\rbrace$, $N^{h}(v_i)=\lbrace v_j\in |V_G|\: |\: A_{ji}\neq 0\rbrace$, $N^{t}(v_i)=\lbrace v_j\in |V_G|\: |\: A_{ij}\neq 0\rbrace$,$N^{u}(v_i)=N^{h}(v_i)\cup N^{t}(v_i)$, $\oplus$  denotes a permutation invariant function and $\psi$, $\varphi$ denote differentiable functions (weaker regularity assumptions can be made). Many popular message passing mechanisms are a particular case of the following one:
\begin{align}\label{gcneq}
    h_{v_i}'=\sigma(\sum_{v_j\in N^{\alpha}(v_i)}c_{f^{\alpha}(i,j)}A_{f^{\alpha}(i,j)}Wh_{v_j}+l_{i}A_{ii}Bh_{v_i})
\end{align}
here $\sigma$ is an activation function, $f^{\alpha}(i,j)=(i,j)$ if $v_j\in N^{t}(v_i)$ and $(j,i)$ if $v_j\in N^{h}(v_i)$, $c_{ij}$, $l_i$ are optional normalization coefficients, $W,B$ are matrices of weights. For digraphs, the choice of $\alpha$ should be thought of as whether a node embedding should be updated by looking at the information of the nodes that are sending to it a signal, by looking at the information of the nodes that are receiving from it a signal or both. These are three different semantics, all justifiable depending on the problem at hand. Two important graph convolutions arise as particular cases of the previous formula: Kipf and Welling's graph convolution for undirected graphs (see \cite{Kipf2016}) and GraphSage convolution (\cite{Sage}) as it appears in the popular package PyTorch Geometric (\cite{pytorch}, \cite{pytorchgeometric}.
Notice that message passing mechanisms as in (\ref{eq1}) are \emph{permutation invariant} in the sense that they do not depend on the ordering given to the set of vertices and only depend on the topology of the underlying graph structure.
We remark that in \cite{Kipf2016} and \cite{Sage} the above convolutions are described only for undirected graphs, but the formulas also make sense for digraphs. In fact the standard implementations used by practitioners are already capable to handle digraphs and are being used also in that context. For example, some papers introducing attention mechanisms (e.g. GAT, see \cite{GAT17}) explicitly introduce this possibility. However, in the digraph case the underlying mathematical theory is somewhat less mature (see \cite{Tong2020} and the reference therein, for the roots of the mathematics behind the directed graph Laplacian the reader is referred to \cite{Bauer2011}).

\subsection{Graph Neural Networks for time series as directed graphs.}
There are many ways to turn time series into digraphs with features. To start with, we introduce two basic examples.\\
\textbf{Example 1}: A multivariate time series $\textbf{x}\in\TS(n,m)\cong\RR^{n\times m}$ can be seen as an unweighted digraph with features as follows. To start with, we consider a set of $n\times m$ nodes $v_{ij}$, with $i=1,...,n$ and $j=1,...,m$.  Then, we create edges  $v_{ij}\rightarrow v_{lk}$ only if $l\geq i$ and the edge is not a self loop (i.e. edges receive information only from the present and the past).
We assign the scalar $\textbf{x}(i)_j$ as feature for each node $v_{ij}$. This construction results in an unweighted digraph with features $(G_\textbf{x}, F_\textbf{x}=(\textbf{x}_{ij}\in\RR))$. One can modify the topology of the graph just constructed. For example, one could create edges $v_{ij}\rightarrow v_{lk}$ only if the edge is not a self-loop, $l\geq i$ and $l-i=0,1$ or if the edge is not a self-loop, $l\geq i$, and $l-i$ is both divisible by a given positive integer $d$ and is smaller than $k\cdot d$ for a given positive integer $k$. This construction results in the directed graph structure pioneered in \cite{BPabstract}): see figure \ref{tgraph}.\\
\textbf{Example 2}: A multivariate time series $\textbf{x}\in\TS(n,m)\cong\RR^{n\times m}$ can be seen as a one dimensional time series with $m$ channels. In this case, the time series can be turned into a digraph with features $(G_\textbf{x}, F_\textbf{x}=(\textbf{x}(i)\in\RR^m))$ by considering a directed graph of $n$ nodes $v_{i}$, $i=1,...,n$ and edges $v_{i}\rightarrow v_{l}$  are added if the edge is not a self-loop and $l\geq i$, $l-i=1$ or and $l-i$ is both divisible by a given positive integer $d$ and is smaller than $k\cdot d$ for a given positive integer $k$. We assign the vector $\textbf{x}(i)\in\RR^m$ as a vector of features for each node $v_{i}$. This completes the construction of the desired directed digraph with features.\par
These examples are of course just a starting point and one can take advantage of further domain knowledge to model the topology of the graph in a more specific way. For instance, one could use auto-correlation functions, usually employed to determine the order of ARMA models (see \cite{TsayTS}), to choose the right value for parameters like $k$ or $d$. As proved in Lemma \ref{1dconv} under certain hypotheses, ordinary convolutions on time series can be seen as transformations between graphs, that is graph convolutions, however the latter evidently carry a very different meaning than ordinary TCNs and can be more flexible. Thus thinking of a time series as a digraph opens to a whole new set of possibilities to be explored. For example, they can be effective as temporal pooling layers when combined with other algorithms or they can leverage the message passing mechanisms that is thought to be more effective to solve the task at hand.

\begin{lem}\label{1dconv}
Consider a convolution $\mathrm{conv1d}:\RR^d\cong\TS(d,1)\rightarrow\TS(d-r,1)\cong\RR^{d-r}$, $(\mathrm{conv1d}(\textbf{x}))_i=\sum_{j=0}^{r-1} K_{j}\textbf{x}_{i+j}$.
Then there exists a weighed digraph $G$, a graph convolution $\mathrm{gconv}:\Feat(G,1)\rightarrow \Feat(G,1)$ and a subgraph $\iota:H\subseteq G$ such that $\Feat(G,1)\cong\TS(d,1)$, $\Feat(H,1)\cong\TS(d-r,1)$ and, under these bijections, $\mathrm{conv1d}$ arise as the map $\iota^\ast\circ\mathrm{gconv}:\Feat(G,1)\rightarrow \Feat(H,1)$.
\end{lem}
\begin{proof}
Define $G$ to be the digraph having $d$ vertices $v_1,...,v_d$ and whose weighted adjacency matrix is given by $A_{ij}= K_{j-i+1}$ if $1\leq j-i+1\leq r$ and zero otherwise. Let $\iota:H\subseteq G$ be its weighed subgraph consisting of the vertices $v_{r},...,v_{d}$. We consider the graph convolution $\mathrm{gconv}:\Feat(G,1)\rightarrow \Feat(G,1)$ arising from the message passing mechanism given by formula (\ref{gcneq}) with $\alpha = h$, $W=1$, $l_i=0$, $\sigma=1$ and $c_{ij}=1$. We define the bijection $\TS(d,1)\cong\Feat(G,1)$ as follows: for each $\textbf{x}\in \TS(d,1)$, $\textbf{x}(i)$ becomes the feature of the node $v_i$ in $G$ (and analogously for $\TS(d-r,1)\cong\Feat(H,1)$).

\end{proof}
The previous lemma can be extended mutatis mutandis also to the case of multivariate time series and contains as a particular case the one of dilated convolutions. The process of learning the weights of a dilated convolution can be thought as the process of learning the weights of the adjacency matrix of a graph.
\begin{rem}\label{Expltimegconv}
Simple Laplacian-based graph convolutions on undirected graphs can be seen as 1-step Euler discretizations of a Heat equation (see fore example \cite{Fioresi2023DeepLA}, \cite{Bronstein2016GeometricDL}). In general, a GNN consisting of a concatenation graph convolutions can be thought as a diffusive process of the information contained in the nodes along the edges: in our context, directed graph convolutions applied to time series digraphs "diffuse" the information through time by updating at each step node features using the information coming from a 'temporal neighbourhood'.
\end{rem}

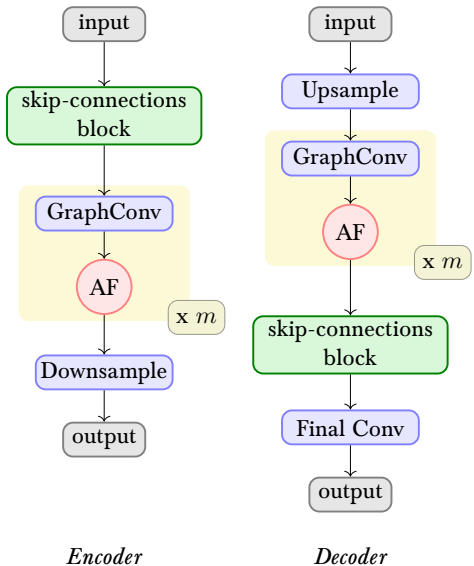
\begin{figure}[h]
\begin{center}
\scalebox{.90}{
\begin{tikzpicture}[scale=.45]
\node at (-4,0) [inout] (input_enc) {input};
\node at (-4,-3) [main] (ly1_enc) {\begin{tabular}{c} skip-connections \\ block \end{tabular}};

\node at (-4,-6.2) [layer] (gconv_enc) {GraphConv};
\node at (-4,-8.6) [bn] (af2_enc) {AF};
\node at (-4,-11.4) [layer] (dn_enc) {Downsample};
\node at (-4,-13.6) [inout] (output_enc) {output};
\node at (-4,-17.4) {\emph{Encoder}};

\draw [->] (input_enc.south) -- (ly1_enc.north);
\draw [->] (ly1_enc.south) -- (gconv_enc.north);
\draw [->] (gconv_enc.south) -- (af2_enc.north);
\draw [->] (af2_enc.south) -- (dn_enc.north);
\draw [->] (dn_enc.south) -- (output_enc.north);

\node at (4,0) [inout] (input_dec) {input};
\node at (4,-2.2) [layer] (up_dec) {Upsample};
\node at (4,-4.4) [layer] (gconv_dec) {GraphConv};
\node at (4,-6.8) [bn] (af1_dec) {AF};
\node at (4,-10.5) [main] (ly1_dec) {\begin{tabular}{c} skip-connections \\ block \end{tabular}};
\node at (4, -13.2) [layer] (final_conv) {Final Conv};
\node at (4,-15.4) [inout] (output_dec) {output};
\node at (4, -17.4) {\emph{Decoder}};

\draw [->] (input_dec.south) -- (up_dec.north);
\draw [->] (up_dec.south) -- (gconv_dec.north);
\draw [->] (gconv_dec.south) -- (af1_dec.north);
\draw [->] (af1_dec.south) -- (ly1_dec.north);
\draw [->] (ly1_dec.south) -- (final_conv.north);
\draw [->] (final_conv.south) -- (output_dec.north);

\begin{pgfonlayer}{background}
\node at (-4,-7.5) [bg] {};
\node at (4,-5.7) [bg] {};
\node at (-1,-9.6) [times] {$\textnormal{x }m$};
\node at (7,-7.8) [times] {$\textnormal{x }m$};
\end{pgfonlayer}
\end{tikzpicture}}
\end{center}
\caption{\footnotesize{The structure of the main building block, both as an encoder and a decoder}}\label{mb}
\end{figure}

\section{Our Models}
We propose two different types of models, both taking advantage of the time-digraph structure described in the previous section: a supervised classifier/regressor and an unsupervised method made of two separate steps, an autoencoder to reconstruct the time series, followed by a clustering algorithm applied to the reconstruction errors.
The core of the algorithm is the same for the two approaches, so we will first focus on this core building block and then proceed to discuss the two models separately.

\subsection{Main Building Block}
The main building block for all the models presented in this paper is a collection of layers that combines TCNs with GNNs. Inspired by what has already been done in the literature, we propose this main building block in two versions: an encoder and a decoder (they are sketched in Figure \ref{mb}(a)). They can be used on their own or put together to construct an autoencoder model.
\par
In the encoder version, the input is first given to a block of either $n$ TCN layers or $n$ GNN layers (we tested primarily Sage convolution-based layers as they appeared more effective after some preliminary tries) with skip connections, following the approach proposed in \cite{Thill2021}. The effectiveness of skip-connections in the model developed in \emph{op.cit} is due to the fact that stacking together the outputs of dilated convolutions with different dilations allows to consider short and long term time dependencies at the same time. Skip connections have been used also in the context of GNN: for example, Sage convolutions (\cite{Sage}) introduce them already in the message passing mechanism. This motivates the introduction of skip connections in GNNs handling digraphs with features coming from time series: in this context they do not only allow to bundle together information coming from "temporal neighbourhoods" of different radii as in the architecture developed in \cite{Thill2021} but they also help to reduce oversmoothing that traditionally curses GNNs architectures (see \cite{Bodnar2022NeuralSD} for a discussion and the references therein).
The skip-connections block is described in figure (\ref{mb}(b)) and works as follows. The input goes through the first TCN/GNN layer, followed by a 1-dimensional convolution and an activation function. We call the dimension of the output of this convolution skip dimension. Then the output is both stored and passed to the next TCN/GNN layer as input and so on. In the architectures we have tested, the TCN/GNN layers are simply dilated 1d convolutions or a single graph convolutions (followed by an activation function), but more involved designs are possible. At the end all the stored tensors are stacked together and passed to a series of $m$ graph convolutions, each one followed by an activation function. We tested the convolutions: GCN (cfr. \cite{Kipf2016}), Sage (cfr. \cite{Sage}), GAT (cfr. \cite{GAT17}).

\begin{figure}[h]
\begin{center}
\scalebox{.90}{
\begin{tikzpicture}[scale=.45]
\node at (-4,0) [inout] (input) {input};
\node at (-4,-2.2) [main] (ly1) {TCN/GCN};
\node at (-4,-4.2) [layer] (conv1) {Conv $1d$};
\node at (-4,-6.5) [bn] (af1) {AF};
\node at (-4,-8.9) [inout] (middle) {$\cdots$};
\node at (-4,-11) [main] (ly2) {TCN/GCN};
\node at (-4,-13) [layer] (conv2) {Conv $1d$};
\node at (-4,-15.2) [bn] (af2) {AF};
\node at (-4,-17.8) [inout] (output) {output};

\draw [decorate, decoration={brace,amplitude=5pt}] (0,-5) -- (0,-13);

\draw [->] (input.south) -- (ly1.north) ;
\draw [->] (ly1.south) -- (conv1.north);
\draw [->] (conv1.south) -- (af1.north);
\draw [->] (af1.south) -- (middle.north);
\draw [rounded corners=5pt,->] (af1.west) --++(-4,0) |- (output.190);
\draw [->] (middle.south) -- (ly2.north);
\draw [rounded corners=5pt,->] (middle.west) --++(-3,0) |- (output.180);
\draw [->] (ly2.south) -- (conv2.north);
\draw [->] (conv2.south) -- (af2.north);
\draw [->] (af2.south) -- (output.north);
\draw [rounded corners=5pt,->] (af2.west) --++(-3,0) |- (output.170);

\begin{pgfonlayer}{background}
\node at (-4,-4.5) [bg3] {};
\node at (-4,-13.2) [bg3] {};
\node at (2.3,-9) [times] {$n\textnormal{ times}$};
\end{pgfonlayer}
\end{tikzpicture}}
\end{center}
\caption{\footnotesize{The structure of the block with skip connections}}\label{skip}
\end{figure}
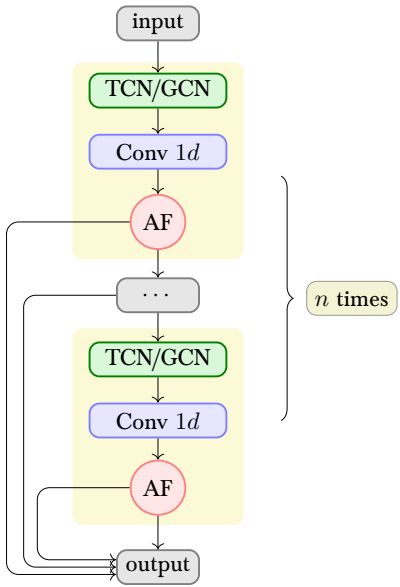

\par
The graph convolutions are defined to encode in the embedding of a given node, at each pass, the information provided by all the nodes in its neighbourhood. Now, looking at how a time-digraph is defined, one sees that in our set up this translates to building the embedding of a given node, that is a data point of the time series, using the information given by the data points that are close in time (short-term memory behaviour) or at a certain distance $d$ away in time (long-term memory behaviour), where $d$ is set in the construction of the graph.
\par
Finally the intermediate output produced by the graph convolutions is given to an optional 1d convolution with kernel size $1$ to adjust the number of channels and then to either an average pooling layer or a max pooling layer that shrinks the temporal dimension of the graph. In other words, if we think about the time-graph as a time window of length $T$, the pooling layers outputs a time window, and therefore a time-graph, of length $T/s$, thus realizing the characteristic \emph{bottleneck} of an autoencoder as described for instance in \cite{Thill2020}, \cite{Thill2021}, \cite{Khandual2021}. We will refer to $s$ as the shrinking factor.
\par
The decoder version changes the order of the blocks we just described, in a symmetric way. It starts with an upsample of the input time series which is then passed to the the graph convolutions followed by the skip-connections block. It terminates with a 1d final convolution with kernel size $1$ that reduces the number of channels, or hidden dimensions, thus giving as output a time series of the same dimensions as the initial input. In the case where the skip-connections block is built with GNN layers, this final convolution can be replaced by a final TCN layer.

\subsection{Regression/Classification}
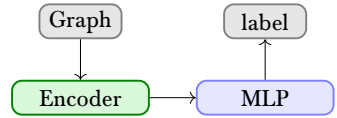
\begin{figure}[h]
\begin{center}
\scalebox{.90}{
\begin{tikzpicture}[scale=.45]
\node at (-3,0) [inout] (input) {Graph};
\node at (-3,-2.5) [main] (mb) {Encoder};
\node at (3,-2.5) [layer] (mlp) {MLP};
\node at (3,0) [inout] (output) {label};

\draw [->] (input.south) -- (mb.north);
\draw [->] (mb.east) -- (mlp.west);
\draw [->] (mlp.north) -- (output.south);

\end{tikzpicture}}
\end{center}
\caption{\footnotesize{The structure of the classifier (or regressor)}}\label{cl}
\end{figure}
The classifier/regressor model uses the main building block described above as an encoder.
The input graph is given to the encoder which predicts a suitable embedding for each of its nodes. At this point the embeddings of the single nodes are combined together into a vector that gives a representation of the whole graph. Recalling that each graph represent a time window, one can think of this first part as a way to extract a small number of features from each time window. These features are then fed to a multi-layer perceptron that outputs a vector of probabilities in the case we use it as a classifier or a value in the case it is used as a regressor instead. As for the way the node embeddings are combined together we explored a few possibilities in the context of classification, the two main options being a flattening layer and a mean pooling layer.
For easiness of notation, from now on we will refer to these models as
TCNGraphClassifier/Regressor if the skip-connections block uses TCN layers,  TGraphClassifier/Regressor if it is built with graph convolutions.
\subsection{Autoencoders for unsupervised anomaly detection}
\begin{figure}[h]
\begin{center}
\scalebox{.90}{
\begin{tikzpicture}[scale=.45]
\node at (-3,0) [inout] (input) {$G_i$};
\node at (-3,-2.5) [main] (enc) {Encoder};
\node at (3,-2.5) [main] (dec) {Decoder};
\node at (3,0) [inout] (output) {$\hat G_i$};

\draw [->] (input.south) -- (enc.north);
\draw [->] (enc.east) -- (dec.west);
\draw [->] (dec.north) -- (output.south);

\end{tikzpicture}}
\end{center}
\caption{\footnotesize{The structure of the autoencoder}}\label{aetcn}
\end{figure}
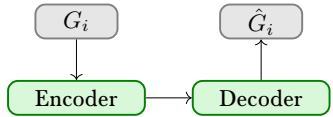
The second architecture we propose is a an autoencoder model, thus it employs two main building blocks, first used as an encoder and then as a decoder and the output is the reconstruction of the given time series represented by the time graph.
In our experiments we use the signal reconstruction obtained with this architecture for anomaly detection purposes. Let us briefly describe our method. The main idea is that the autoencoder model provides a good reconstruction of the input time series when the signal is normal and worst reconstructions on time windows where an anomaly appears (again we refer to \cite{Thill2021} among others for a similar approach to anomaly detection), as it is constructed to remove noise from a signal.
Thus, once we have the reconstructed times series, we compute both the Root Mean Square Error and the Mahalanobis score (see \cite{Thill2021} for more details in a similar context), for each given time window, with respect to the original time series. In the case one has to deal with more than one time series bundled together in the time-digraph, there are simple methods to get a single score for each time window. Now we can treat these two measures of the reconstruction error as features of the time windows and use an unsupervised clustering algorithm (we tested both Kmeans and DBscan), to assign a binary label to each window, based on the cluster they fall into (see Section \ref{unsup} for more details). \par
This approach gives a completely unsupervised method to handle anomaly detection of one or more time series.
Again, from now on we will refer to these models as:
TCNGraphAE if the skip-connections block uses TCN layers, TGraphAE if it is built with graph convolutions.
and TGraphMixedAE if the encoder uses graph convolutions and the Decoder uses TCN layers. If the skip connections blocks consist only of dilated convolutions and we do not have a final graph convolution to filter the signal, we obtain an TCN autoencoder/Classifier with a structure similar to the one described in \cite{Thill2021}. We call this latter models TCNAE and TCNClassifier. We regard these models as state of the art models in this context and we use them as a benchmark.

\section{Experiments}
For our experiments we used a database made of ECG signals. These signals have been recorded with a ButterfLive medical device at a sampling rate of 512 Hz. A lowband Butterworth filter at 48 Hz was applied to each signal. Then every 5 second long piece of signal was manually labeled according to readability of the signal: label 3 was given to good quality signals, label 2 was given to medium quality signals and label 1 was given to low quality/unreadable signals. In total we had a database made of 10590 5-second-long sequences.
\par
We turned the problem into a binary classification: label 0 was assigned to signals having label=1 and label 1 was assigned to signals having label=2,3. A Train/Valid/Test split was performed on the database with weights 0.3/0.35/0.35. Train, Valid and Test sets are made of signals coming from different recordings and the signals having label 0 are approximately the 18\% of each set. For the final evaluation of our models, we run the models for $10$ times, then, for each score, the best and the worst results were removed and the mean and standard deviation of the remaining $8$ runs were computed and are reported in Tables \ref{tab1}, \ref{tab2}. 

\begin{table*}[!t]
\begin{center}
\footnotesize
\makebox[1 \textwidth][c]{
\begin{tabular}{c |c c c | c c c}
\hline
\multirow{2}{*}{\textbf{Model}} & \multicolumn{3}{c|}{Positive class = label $1$} & \multicolumn{3}{c}{Positive class = label $0$}\\ 
& \textbf{Precision} & \textbf{Recall} & \textbf{Accuracy} & \textbf{Precision} & \textbf{Recall} & \textbf{Accuracy} \\
\hline
TGraphClassifier & $0.965 \pm 0.002$ & $0.991 \pm 0.002$ & $0.962 \pm 0.003$ & $0.941 \pm 0.012$ & $0.806 \pm 0.011$ & $0.962 \pm 0.003$ \\
TCNGraphClassifier & $0.939 \pm 0.013$ & $0.988 \pm 0.006$ & $0.936 \pm 0.010$ & $0.912 \pm 0.044$ & $0.653 \pm 0.083$ & $0.936 \pm 0.010$\\
TCNClassifier & $\textcolor{purple}{0.975} \pm 0.003$ & $\textcolor{purple}{0.994} \pm 0.002$ & $\textcolor{purple}{0.973} \pm 0.003$ & $\textcolor{purple}{0.962} \pm 0.011$ & $\textcolor{purple}{0.863} \pm 0.017$ & $\textcolor{purple}{0.973} \pm 0.003$ \\
\hline
\end{tabular}}
\end{center}\caption{\footnotesize{Results of the classifiers. Best scores are colored in purple.}}\label{tab1}
\end{table*}
\begin{table*}[!t]
\begin{center}
\footnotesize
\makebox[1 \textwidth][c]{
\begin{tabular}{c | c c c | c c c}
\hline
\multirow{2}{*}{\textbf{Model}} & \multicolumn{3}{c|}{Positive class = label $1$} & \multicolumn{3}{c}{Positive class = label $0$} \\ 
& \textbf{Precision} & \textbf{Recall} & \textbf{Accuracy} & \textbf{Precision} & \textbf{Recall} & \textbf{Accuracy} \\
\hline
\multicolumn{7}{l}{\emph{Kmeans, approach A}}\\
\hline
TGraphMixedAE & $\textcolor{purple}{0.973} \pm 0.006$ & $0.974 \pm 0.011$ & $0.955 \pm 0.006$ & $0.854 \pm 0.051$ & $\textcolor{purple}{0.847} \pm 0.035$ & $0.955 \pm 0.006$\\
TGraphAE & $0.960\pm 0.007$ & $0.993\pm 0.006$ & $0.959\pm 0.003$ & $0.952\pm 0.040$ & $0.765\pm 0.044$ & $0.959 \pm 0.003$ \\
TCNGraphAE1 & $\textcolor{blue}{0.967}\pm 0.003$ & $\textcolor{blue}{0.998}\pm 0.002$ & $\textcolor{purple}{0.968}\pm 0.002$ & $\textcolor{blue}{0.985}\pm 0.014$ & $\textcolor{blue}{0.806}\pm 0.016$  & $\textcolor{purple}{0.969}\pm 0.002$ \\
TCNGraphAE2  & $0.965\pm 0.002$ & $0.997 \pm 0.001$ & $\textcolor{blue}{0.966}\pm 0.002$ & $0.976\pm 0.007$ & $0.796\pm 0.012$ & $\textcolor{blue}{0.966}\pm 0.002$ \\
TCNAE1  & $0.966\pm 0.012$ & $0.995\pm 0.005$ & $0.964\pm 0.009$ & $0.966 \pm 0.032$ & $0.798\pm 0.074$ & $0.964\pm 0.009$ \\
TCNAE2  & $0.949\pm 0.005$ & $\textcolor{purple}{0.999}\pm 0.001$ & $0.951\pm 0.004$ & $\textcolor{purple}{0.995} \pm 0.006$ & $0.692\pm 0.031$ & $0.954\pm 0.004$ \\
\hline
\multicolumn{7}{l}{\emph{Dbscan, approach B}}\\
\hline
TGraphMixedAE & $\textcolor{purple}{0.984} \pm 0.005$ & $0.944 \pm 0.012$ & $0.939 \pm 0.007$ & $0.745 \pm 0.037$ & $\textcolor{purple}{0.909} \pm 0.028$ & $0.939 \pm 0.007$ \\
TGraphAE  & $0.968 \pm 0.006$ & $\textcolor{blue}{0.989}\pm 0.006$ & $0.962\pm 0.002$ & $\textcolor{blue}{0.933}\pm 0.034$ & $0.813\pm 0.038$ & $0.962\pm 0.002$ \\
TCNGraphAE1  & $0.971\pm 0.004$ & $\textcolor{purple}{0.991} \pm 0.005$ & $\textcolor{blue}{0.966}\pm 0.003$ & $\textcolor{purple}{0.940}\pm 0.028$ & $0.829\pm 0.022$  & $\textcolor{blue}{0.966}\pm 0.003$ \\
TCNGraphAE2  & $\textcolor{blue}{0.979}\pm 0.007$ & $0.985 \pm 0.006$ & $\textcolor{purple}{0.967}\pm 0.001$ & $0.913\pm 0.031$ & $\textcolor{blue}{0.877}\pm 0.043$  & $\textcolor{purple}{0.967}\pm 0.001$ \\
TCNAE1   & $0.971 \pm 0.007$ & $0.985\pm 0.011$ & $0.962 \pm 0.007$ & $0.913 \pm 0.057$ & $0.833 \pm 0.043$ & $0.962\pm 0.007$ \\
TCNAE2  & $0.973\pm 0.006$ & $0.988\pm 0.005$ & $\textcolor{blue}{0.966}\pm 0.002$ & $0.925 \pm 0.027$ & $0.846\pm 0.039$ & $\textcolor{blue}{0.966}\pm 0.002$ \\
\hline
\end{tabular}}
\end{center}\caption{\footnotesize{Results of the autoencoder algorithms. Best scores are colored in purple and the second best in blue.}}\label{tab2}
\end{table*}
\subsection{Supervised Classification}
To test how graph convolutions perform for our classification problem using a supervised method, we applied a convolution smoother of window 20 to our dataset, and then we performed a downsample of ratio 4 (i.e. we kept one point every 4). We subdivided each Train/Valid/Test set in non overlapping slices of 5 seconds and we applied a min-max scaler to each sequence independently. \par
Each 5 second long time series $\textbf{x}\in\TS(640,1)$ was given a simple directed graph structure as in Example 2, consisting of one node per signal's point (resulting in 640 nodes). We call $k\cdot d$ the \emph{lookback window}. We used 128 as our lookback window (1 second) and we set $d$ to be equal to $4$. We used the Adam optimizer to train all our models. Results are displayed in Table \ref{tab1}. Further details about the models used are contained in the Appendix. For the graph convolutions involved in model TCNGraphClassifier the underlying message passings used $\alpha = t$ as in Formula \ref{gcneq}: this results in the time dependencies to be read in the reversed direction by these layers.

\begin{figure*}[t]
\makebox[1 \textwidth][c]{
\includegraphics[scale=0.45]{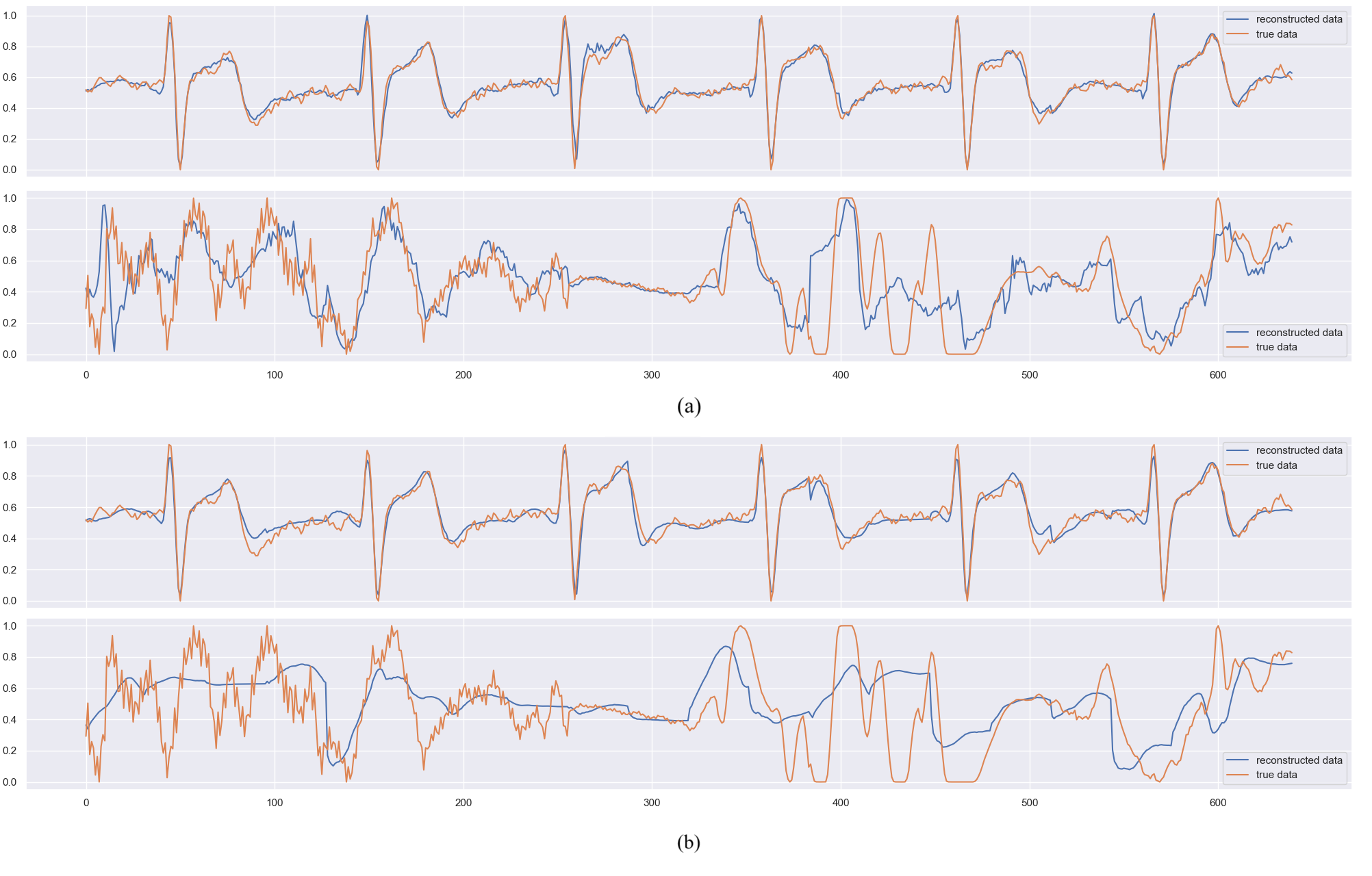}}
\caption{\footnotesize{The reconstructed signals with TGraphAE (a) and TCNAE1 (b). For each subfigure, the top image represents the reconstruction of a good signal (label $1$) and the bottom one is the reconstruction of a bad signal (label $0$).}}\label{figsignal}
\end{figure*}

\subsection{Unsupervised Anomaly Detection with Autoencoders}\label{unsup}
For our unsupervised experiments, the data was preprocessed and prepared as in the supervised case with the only difference that the signals were divided into pieces of one second and that the lookback window was set to 25. Then, we trained the autoencoder to reconstruct these 1 second long signals. $d$ was set to be either $4$ or $8$ depending on the model. \par
We used the following training procedure: first each model has been trained on the train set for $50$-$100$ epochs. Then the worst reconstructed $20$\% of signals was discarded. This decreased the percentage of unreadable data in the training set from approximately $18$\% to $6$\% for each model. Each model was then retrained from scratch for $150$-$325$ epochs on the refined training set. The rationale of this choice is that, for autoencoders to be used effectively as anomaly detectors, the 'anomalies' should be as few as possible to prevent overfitting the anomalies (see \cite{Thill2021}, where signals with too many anomalies were discarded). We found that using an autoencoder trained for less epochs can be employed effectively to reduce the proportion of anomalies in the training set. The trained models were then used to compute, for each signal in the Valid set, the reconstruction loss and the Mahalanobis scores as in \cite{Thill2021}. 
Both the resulting scores were then averaged and normalized to provide one mean reconstruction error and one mean Mahalanobis score for each labeled 5 second slice of signal. We thus obtained a set of pairs of scores, one for each 5 second long signal in the Valid set: we will refer to it as the errors Valid set. \par
\begin{figure*}[t]
\begin{subfigure}[b]{0.245\textwidth}
\includegraphics[scale=0.18]{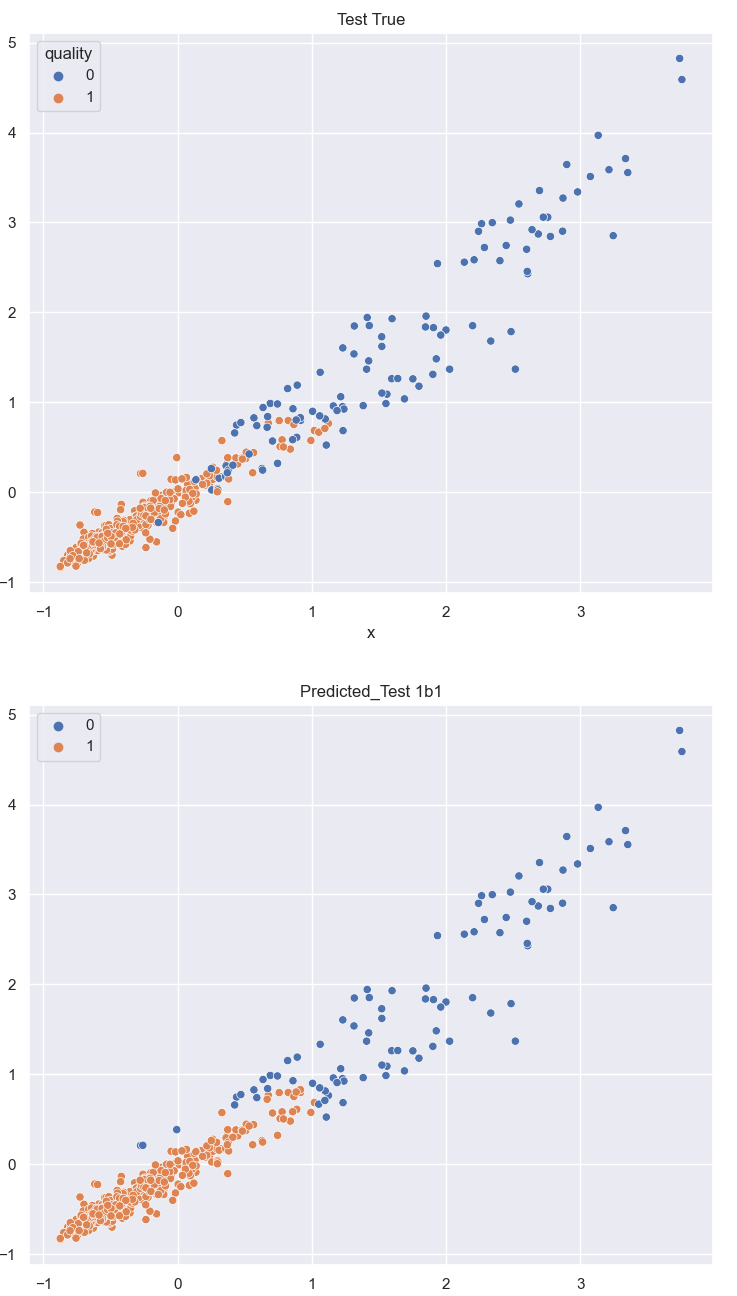}
\caption{}
\end{subfigure}
\begin{subfigure}[b]{0.245\textwidth}
\includegraphics[scale=0.18]{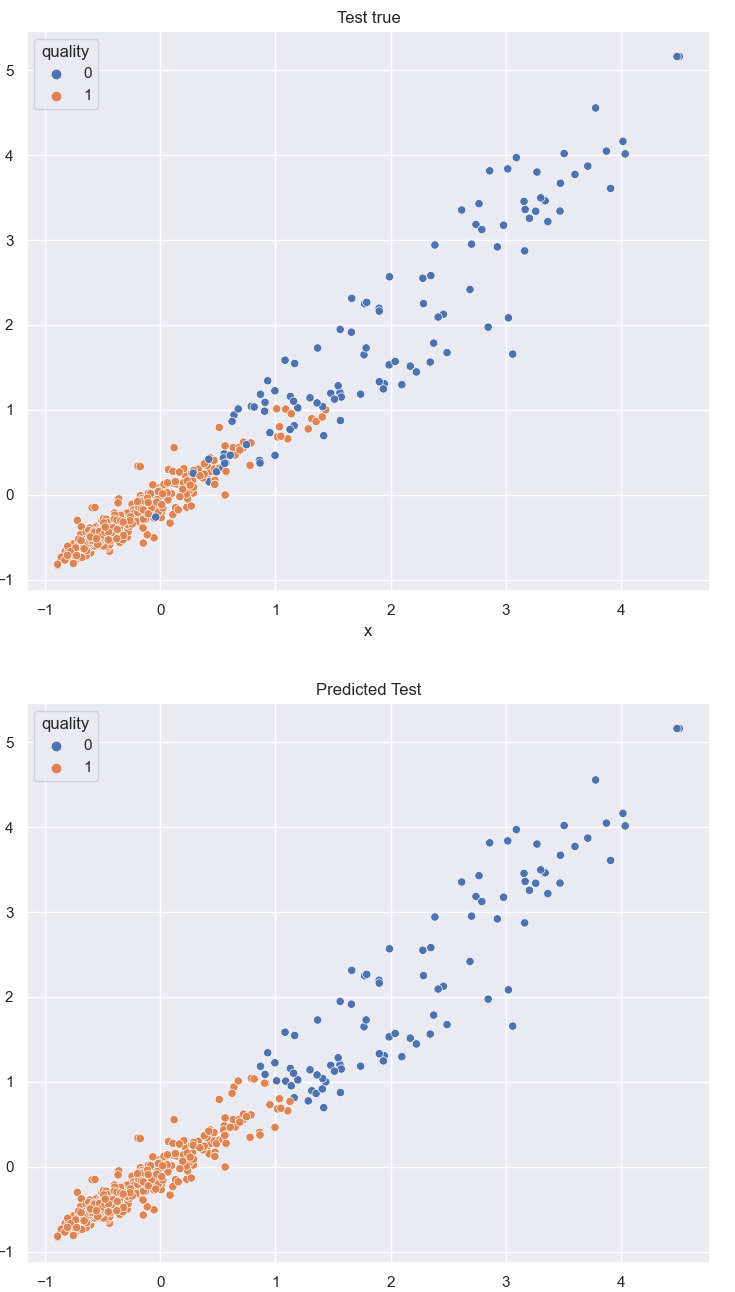}
\caption{}
\end{subfigure}
\begin{subfigure}[b]{0.245\textwidth}
\includegraphics[scale=0.18]{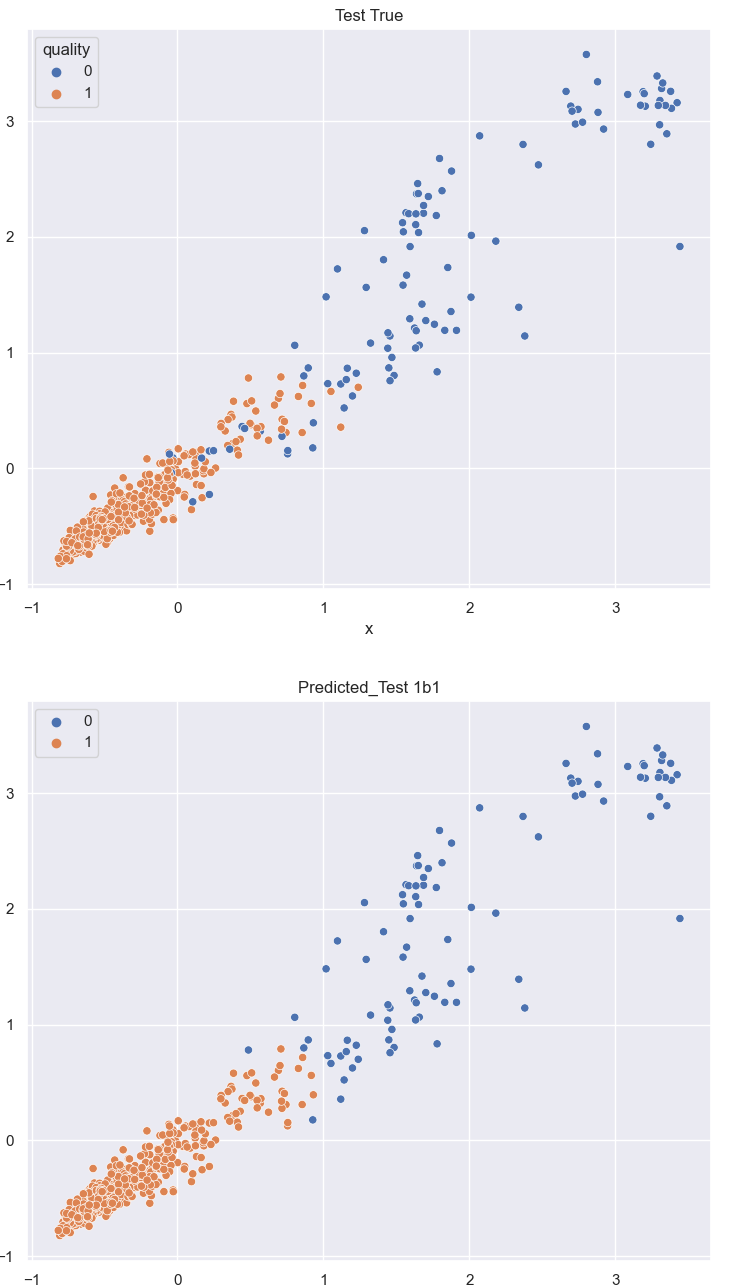}
\caption{}
\end{subfigure}
\begin{subfigure}[b]{0.245\textwidth}
\includegraphics[scale=0.18]{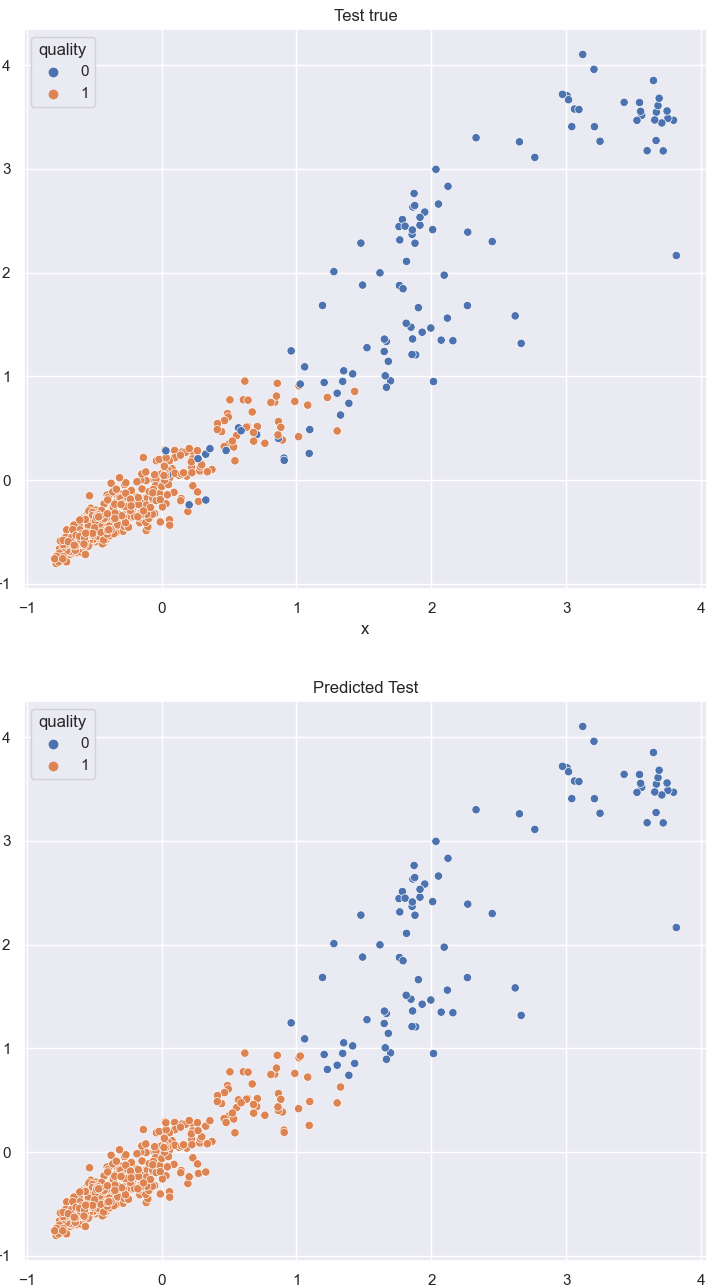}
\caption{}
\end{subfigure}
\caption{\footnotesize{The clusters obtained with TGraphAE and TCNAE1, as follows: (a) TGraphAE-dbscan, (b) TGraphAE-kmeans, (c) TCNAE1-dbscan, (d) TCNAE1-kmeans. For each subfigure, in the top image the points are colored based on their true label and in the one on the bottom they are colored based on the predicted cluster (label $1$ in orange and label $0$ in blue).}}
\label{figclust}\end{figure*}

We used the error Valid set as a feature space to train two unsupervised cluster algorithms: Kmeans and DBscan. For dbscan we set the minimum number of points in a cluster to be equal to $4$, as customary for $2$ dimensional data, and we used as epsilon the mean of the distances of the points plus $2$ times their standard deviation. 
For both, to get the final labels on the Test set, we used two different techniques. One option we considered is to obtain the final labels for the Test set repeating exactly the procedure used for the Valid set (approach A). The second technique we used goes as follows: first, we trained an SVM classifier on the errors Valid set, labeled using the clustering provided by the unsupervised method. Then, we obtained an errors Test set applying the procedure described above for the Valid set, but using the normalizers fitted on the errors Valid set. Finally, we used the trained SVC to predict the labels of the Test signals (approach B). 
We report the results in table \ref{tab2}, while the clusters obtained for models TGraphAE and TCNAE1 are displayed in Figure \ref{figclust}. \par
The signals reconstructed by these models are displayed in Figure \ref{figsignal}. Both models reconstruct good signals in a comparable way and fail to properly reconstruct the bad signals, as expected, despite their small number of parameters. Notice that these methods are fully unsupervised and do not require the use of even a few labeled samples. As in the supervised setting, the graph convolutions involved in models TGraphMixedAE, TCNGraphAE1 and TCNGraphAE1 use $\alpha = t$ in their underlying message passings as in Formula \ref{gcneq}. As a consequence, in these models, time dependencies were read in the right direction by the TCNs and in the reversed one by the graph convolutions, resulting in parameter-efficient "bidirectional" structures.

\subsection{Discussion}
In the case of the supervised classification, a TCN classifier without the use of graph convolutions proved to be the best performing one. This is probably due to the effect of the final flattening layer that may provide the best mechanism in this context to link the encoder to the final MLP. The graph based model had a worse performance but achieved its best using a mean pooling mechanism, as it can be expected. However the graph based classifier obtained good results and had less than half of the parameters than the TCN classifier (see Table \ref{tab3}), thus exhibiting the more expressive power of the graph convolutions. \par
In the case of the unsupervised classification, the best performing models were on average the TCN based ones where graph convolutions were added right before and after the bottleneck. This gives a good indication of the fact that graph convolutions applied to digraphs with features can serve as good layers to filter signal coming from different layers. Also the second best performing model overall consists of a Graph based encoder and a TCN decoder, strengthening the hypothesis that graph convolution can be used to improve the effectiveness of ordinary models or to reduce considerably the number of parameters of state of the art architecture without decreasing too much their performance. It has to be noted that in the best performing models, the message passing mechanisms of the graph convolutions were particular cases of formula \ref{gcneq} with $\alpha = t$: as a consequence, these layers learn time dependencies as if the features learnt by the first part of the encoder where reversed in time. Therefore two types of time dependencies were combined in the same algorithm in a parameter-efficient way, mimicking the behaviour of other 'bidirectional' models (such as bidirectional LSTMs). \par
Summing up, GNN applied to digraphs with features coming from time series showed their effectiveness in improving established algorithms and also their potential to replace them. Moreover, effective fully unsupervised pipelines can be devised to solve anomaly detection and quality recognition problems using the models described in this paper. We plan to continue the study of GNNs applied to time digraphs with features in the context of multivariate time series, constructing more complex time digraph structures and using more capable message passing mechanisms.

\subsection{Reproducibility Statement}
The experiments described in this paper are reproducible with additional details provided in Appendix \ref{hyp}. This section includes a list of all the hyperparameters used to train our models (see Table \ref{tab3}) and more details on the algorithms' implementation.
\subsection{Acknowledgements}
The authors wish to thank Prof. Rita Fioresi for many stimulating discussions on the content of this paper and A. Arigliano, G. Faglioni and A. Malagoli of VST for providing them with the labeled database used for this work. \par
A. Simonetti wishes to thank professor J. Evans for his continuous support.  
The research of F. Zanchetta was supported by Gnsaga-Indam, by COST Action CaLISTA CA21109, HORIZON-MSCA-2022-SE-01-01 CaLIGOLA, MNESYS PE, GNSAGA and has been carried out under a research contract cofounded by the European Union and PON Ricerca e Innovazione 2014-2020 as in the art. 24, comma 3, lett. a), of the Legge 30 dicembre 2010, n. 240 e s.m.i. and of D.M. 10 agosto 2021 n. 1062.

\appendix
\begin{table*}[t!]
\begin{center}
\footnotesize
\makebox[1 \textwidth][c]{
\begin{tabular}{c | c c c c c c c}
\multicolumn{8}{c}{Supervised Models}\\
\hline
\textbf{Model}  & \textbf{Num Channels} & \textbf{Skip dims} & \textbf{GConv type} & \textbf{Pool type} & \textbf{MLP dims} & \textbf{Shrink} & \textbf{Params} \\
\hline
TGraphClassifier & $[32]*4,16$ & $[16]*4$ & Sage & Mean Pool & - & $16$ & $8k$\\
TCNGraphClassifier & $[32]*7,2$ & $[16]*7$ & Sage & Flattening & - & $16$ & $30k$ \\
TCNClassifier & $[32]*4$ & $[16]*4$ & - & Flattening & $[30,30]$ & $16$ & $20k$\\
\hline
\multicolumn{8}{c}{}\\

\multicolumn{8}{c}{Unsupervised Models}\\
\hline
\textbf{Model}  & \textbf{Num Channels} & \textbf{Skip dims} & \textbf{GConv type} & \textbf{Downsample} & \textbf{Upsample} & \textbf{Shrink} & \textbf{Params} \\
\hline
TGraphMixedAE & $[64]*7,2$ & $[32]*7$ & Sage & Graph & Nearest & $16$ & $145k$\\
TGraphAE & $[64]*4,2$ & $[32]*4$ & GAT2H & Average & Nearest & $16$ & $44k$\\
TCNGraphAE1 & $[32]*3,2$ & $[16]*3$ & GAT2H, $[100]$ & Graph & Nearest  & $32$ & $33k$\\
TCNGraphAE2 & $[64]*7,4$ & $[32]*7$ & GAT2H, $[100]$ & Graph & Nearest  & $32$ & $258k$\\
TCNAE1  & $[32]*3,2$ & $[16]*3$ & - & Max & Nearest & $32$ & $19k$\\
TCNAE2  & $[64]*7,4$ & $[32]*7$ & - & Average & Nearest & $32$ & $208k$\\
\hline
\end{tabular}}
\end{center}\caption{\footnotesize{Hyperparameters of the algorithms}}\label{tab3}
\end{table*}
\section{Models' hyperparameters and details}\label{hyp}
The specific hyperparameters of the models described in the previous sections are listed in table \ref{tab3}. Here is a description of this table, to understand all the names and abbreviations there appearing. \par
\emph{Num Channels} gives the number of layers used in the skip connections block, together with the output dimension of each layer in the form of a list; the number after the comma indicates the channel dimension of the signal in the bottleneck resulting after the application of the (1D convolution) graph convolution at the end of the encoder. \par
\emph{Skip dims} gives the list of the skip dimensions as described in Section 5.1. \par
\emph{GConv type} gives the information on the graph convolution that follows the skip connections block, if present: it gives the type of convolution used and the list of its hidden dimensions - in case the convolution is a GAT layer, it also specifies the number of heads chosen (GAT2H, for instance, means that $2$ heads were selected). \par
As for \emph{Pool type}, \emph{Downsample}, \emph{Upsample}, we indicate the type of layer used referring to their standard notation (see for example pytorch geometric libraries).\par
Finally for dilated convolutions, we used a kernel size of $7$ for autoencoders models and $8$ for the supervised models. When skip connections blocks consist of a sequence of dilated convolutions, we used increasing dilations $2^0,2^1,2^2,...,2^n$ where $n$ is the number of convolutions appearing in the considered block. We used SiLU activation functions and we employed both batch normalization and dropout between layers. In model TGraphClassifier, a final 1D dilated convolution with dilation $1$ was applied after the decoder.

\bibstyle{plain}
\bibliography{main_new}

\end{document}